\documentclass[sigconf,natbib]{acmart}

\AtBeginDocument{%
  \providecommand\BibTeX{{%
    \normalfont B\kern-0.5em{\scshape i\kern-0.25em b}\kern-0.8em\TeX}}}

\setcopyright{acmcopyright}
\copyrightyear{2018}
\acmYear{2018}
\acmDOI{XXXXXXX.XXXXXXX}

\acmConference[Conference acronym 'XX]{Make sure to enter the correct
  conference title from your rights confirmation emai}{June 03--05,
  2018}{Woodstock, NY}
\acmPrice{15.00}
\acmISBN{978-1-4503-XXXX-X/18/06}

\usepackage{colortbl,xcolor}

\AtBeginDocument{%
 \providecommand\BibTeX{{%
  Bib\TeX}}}

\copyrightyear{2023}
\acmYear{2023}
\setcopyright{rightsretained}
\acmConference[SIGIR '23]{Proceedings of the 46th International ACM SIGIR Conference on Research and Development in Information Retrieval}{July 23--27, 2023}{Taipei, Taiwan}
\acmBooktitle{Proceedings of the 46th International ACM SIGIR Conference on Research and Development in Information Retrieval (SIGIR '23), July 23--27, 2023, Taipei, Taiwan}\acmDOI{10.1145/3539618.3591957}
\acmISBN{978-1-4503-9408-6/23/07}

\makeatletter
\gdef\@copyrightpermission{
  \begin{minipage}{0.3\columnwidth}
   \href{https://creativecommons.org/licenses/by/4.0/}{\includegraphics[width=0.90\textwidth]{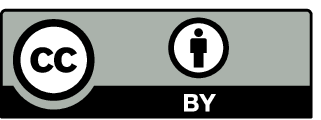}}
  \end{minipage}\hfill
  \begin{minipage}{0.7\columnwidth}
   \href{https://creativecommons.org/licenses/by/4.0/}{This work is licensed under a Creative Commons Attribution International 4.0 License.}
  \end{minipage}
  \vspace{5pt}
}
\makeatother

\begin{document}

\title{BioAug: Conditional Generation based Data Augmentation for Low-Resource Biomedical NER}

\author{Sreyan Ghosh$^{\star}$}
\affiliation{%
  \institution{University of Maryland, College Park}
  \city{College Park}
  \state{MD}
  \country{USA}
}
\email{sreyang@umd.edu}

\author{Utkarsh Tyagi$^{\star}$}
\affiliation{%
  \institution{University of Maryland, College Park}
  \city{College Park}
  \state{MD}
  \country{USA}
}
\email{utkarsh4430@gmail.com}

\author{Sonal Kumar$^{\star}$}
  \affiliation{%
  \institution{University of Maryland, College Park}
  \city{College Park}
  \state{MD}
  \country{USA}
}
\email{skbrahee@gmail.com}

\author{Dinesh Manocha}
  \affiliation{%
  \institution{University of Maryland, College Park}
  \city{College Park}
  \state{MD}
  \country{USA}
\thanks{\hspace*{-1mm}$^{\star}$Equal technical contribution}}
\email{dmanocha@umd.edu}

\renewcommand{\shortauthors}{Sreyan Ghosh, Utkarsh Tyagi, Sonal Kumar, \& Dinesh Manocha}

\newcommand\blfootnote[1]{%
  \begingroup
  \renewcommand\thefootnote{}\footnote{#1}%
  \addtocounter{footnote}{-1}%
  \endgroup
}

\begin{abstract}
Biomedical Named Entity Recognition (BioNER) is the fundamental task of identifying named entities from biomedical text. However, BioNER suffers from severe data scarcity and lacks high-quality labeled data due to the highly specialized and expert knowledge required for annotation. Though data augmentation has shown to be highly effective for low-resource NER in general, existing data augmentation techniques fail to produce factual and diverse augmentations for BioNER. In this paper, we present BioAug, a novel data augmentation framework for low-resource BioNER. BioAug, built on BART, is trained to solve a novel text reconstruction task based on \emph{selective masking} and \emph{knowledge augmentation}. Post training, we perform conditional generation and generate diverse augmentations conditioning BioAug on selectively corrupted text similar to the training stage. We demonstrate the effectiveness of BioAug on 5 benchmark BioNER datasets and show that BioAug outperforms all our baselines by a significant margin (1.5\%-21.5\% absolute improvement) and is able to generate augmentations that are both more factual and diverse. Code: https://github.com/Sreyan88/BioAug.

\end{abstract}


%






\maketitle


\section{Introduction}
Biomedical Named Entity Recognition (BioNER) is the task of detecting named entities (NEs) from unstructured biomedical text (e.g., chemicals, diseases, genes, etc.), which are further utilized in many important downstream tasks (e.g., classifying drug-drug interaction). Compared to NER in the general domain (e.g., news), BioNER is much more challenging due to the syntactically complex nature of NEs \cite{liu2015drug}, large variations in NE mentions \cite{jia-etal-2019-cross,kim2019neural}, and the rapidly emerging nature of NEs in biomedical literature \cite{luo2018attention}. Biomedical NLP, however, suffers from acute data scarcity  due to the high cost involved in leveraging extensive domain knowledge for annotation \cite{wang2019clinical}.  Our experiments show that the performance of the state-of-the-art NER system \cite{zhou2021learning}, when compared to CoNLL 2003 \cite{sang2003introduction} (news domain), drops by 35\% and 21\% respectively, on low- and high-resource settings (100 and 2000 gold training samples) when evaluated on the BC2GM dataset \cite{smith2008overview} (biomedical domain).

\begin{figure*}[t]
\centering
\includegraphics[width=1.8\columnwidth]{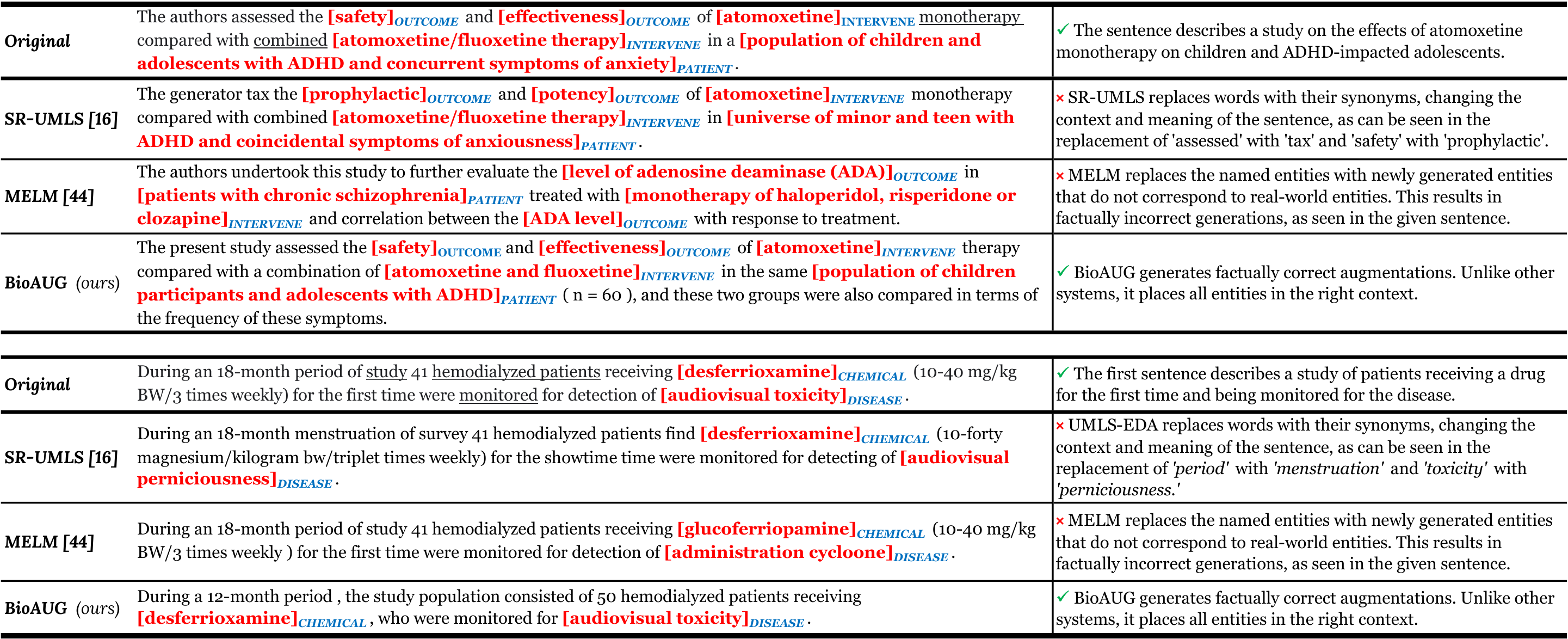}
    \caption{\small{Augmentation examples (left) on the EBMNLP dataset (top) and BC5DR dataset (bottom) and their explanations (right). Words in \textcolor{red}{red} are Named Entities, and words \underline{underlined} are \emph{keywords}. BioAug generates augmentations that are both more factual and diverse.}}
    \label{fig:conll}
\end{figure*}

Data augmentation has proven to be an effective solution for low-resource BioNER \cite{kang2021umls,phan2022simple} and low-resource NER in general \cite{ding2020daga,zhou-etal-2022-melm}. The primary aim is to find \emph{n} transforms of the original training samples that can be added to the original training data to improve learning. However, this area is relatively under-explored due to the difficulty of the task - generic augmentation systems which work for classification tasks suffer from token-label misalignment issues when applied to NER, as the labels in a classification dataset are less sensitive to the sequence of text than the semantics \cite{kang2021umls,zhou-etal-2022-melm}. Adding to this, existing NER data augmentations based on token replacement \cite{dai2020analysis} or Pre-trained Language Models (PLMs) \cite{zhou-etal-2022-melm,liu2021mulda} often fail to generate diverse, and factual\footnote{We call a generated augmentation counterfactual if it has a context-entity mismatch, e.g., a NE belonging to a skin disease is placed in the context of leg pain, Context not contributing to defining the NEs can take any form and promote diversity.} augmentations - biomedical NEs are linguistically complex than NEs in the general domain and methods like synonym replacement \cite{dai2020analysis} often lead to counterfactual augmentations. Moreover, PLMs suffer from inadequacy of knowledge about these NEs due to their inability to generalize over rapidly increasing biomedical concepts (examples in Figure \ref{fig:conll}). We argue that both these metrics are extremely important for data augmentation in BioNER. Factual augmentations promote a model to store factual knowledge in their parameters, which is crucial for the knowledge-intensive nature of BioNLP \cite{zhang2021smedbert}. On the other hand, diversity in augmentations has proven to be a key metric benefiting low-resource NLP \cite{chen2021empirical}.
\vspace{1mm}

\begin{figure*}[t]
\centering
\includegraphics[width=2\columnwidth]{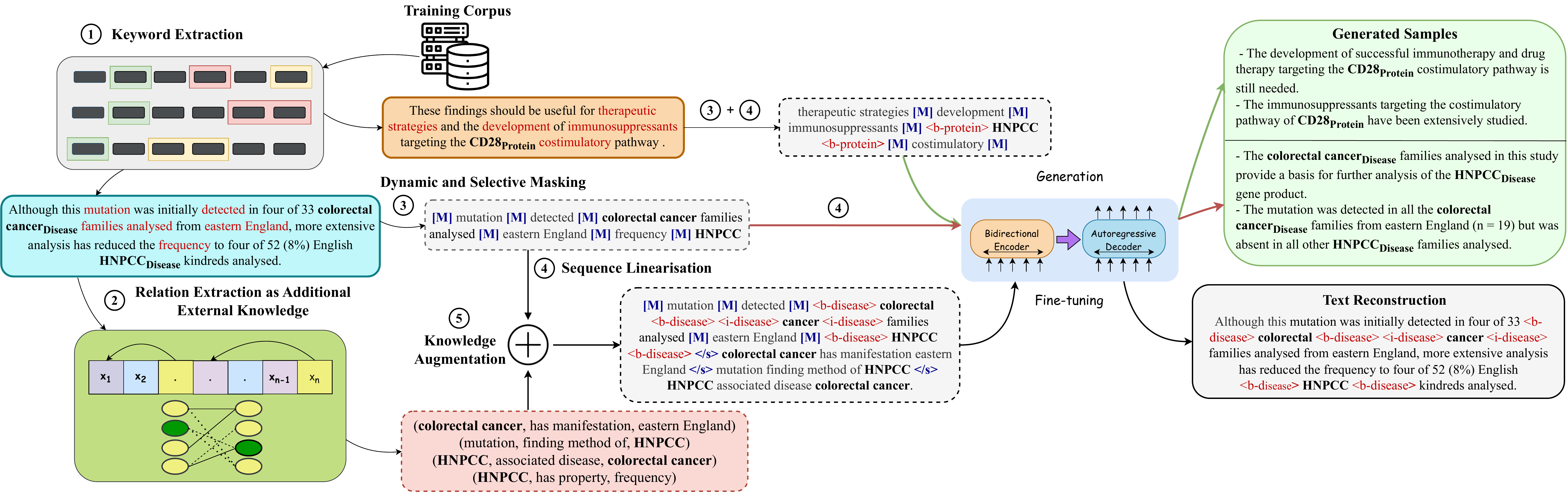}
\caption{\small Overview of \textbf{BioAug:} BioAug builds on BioBART \cite{yuan-etal-2022-biobart} and follows a 5-step sentence corruption process for fine-tuning: \textcircled{\raisebox{-0.9pt}{1}} \textbf{Keyword Extraction:} We use a pre-trained entity extraction model to extract keywords from a sentence which we call \emph{keywords}. \textcircled{\raisebox{-0.9pt}{2}} \textbf{Relation Extraction:} We then use another pre-trained model to extract entity-entity, entity-NE, and NE-NE relations. \textcircled{\raisebox{-0.9pt}{3}} \textbf{Dynamic and Selective Masking:} We randomly mask $e\%$ of the keywords at each iteration. \textcircled{\raisebox{-0.9pt}{4}} \textbf{Sequence Linearization:} We perform linearization and add label tokens before and after each NE token. \textcircled{\raisebox{-0.9pt}{5}} \textbf{Knowledge Augmentation:} We verbalize extracted relations and concatenate them with the masked sentence as knowledge triples. Post fine-tuning, we corrupt a sentence following steps \textcircled{\raisebox{-0.9pt}{1}} - \textcircled{\raisebox{-0.9pt}{4}} and feed it to BioAug to generate diverse and factual augmentations. } 
\label{fig:BioAug}
\end{figure*}

{\noindent \textbf{Main Results.}} In this paper, we present BioAug, a novel data augmentation framework for low-resource biomedical NER. BioAug builds on the core idea of conditional generation and has 2 main steps, fine-tuning, and generation: (1) In the fine-tuning step, we use a pre-trained transformer-based auto-regressive Encoder-Decoder model (BioBart \cite{yuan-etal-2022-biobart} in our case) and fine-tune it on a novel text reconstruction or denoising task. Different from usual denoising-based PLM pre-training objectives \cite{devlin2018bert,lewis2019bart}, which randomly mask tokens in the sentence to corrupt it, we perform \emph{selective masking} and mask all tokens in the sentence except the ones belonging to the NEs and a small percentage of \emph{keywords}. These \emph{keywords} are other important tokens that belong to other entities extracted from an entity extraction model beyond the gold-annotated NEs and are selected randomly for each sentence at each training epoch. Additionally, to improve biomedical text understanding with high-quality knowledge facts, which are difficult to learn from just raw text, we provide the model with contextually relevant external knowledge and train it to use this knowledge. To be precise, in the fine-tuning stage, together with the corrupted sentence, we concatenate knowledge triples in the form of inter-entity relations, which are obtained from a relation extraction (RE) model trained on a large-scale biomedical RE dataset. We explain our rationale behind every step in Section \ref{sub:finetuning}. (2) Post fine-tuning, to generate augmentations, we feed randomly corrupted training data to the model and ask the model to reconstruct sentences from it. Fig. \ref{fig:BioAug} shows a clear pictorial representation of BioAug fine-tuning and generation steps. In practice, BioAug generates augmentations that are both diverse and factual and outperforms all our baselines on benchmark BioNER datasets by a significant margin with absolute improvements in the range of 1.5\% - 21.5\%.


\section{Related Work}
{\noindent \textbf{Bio-medical NER.}} Bio-medical NER is an emerging field of research with numerous systems proposed in the past decade \cite{9746482,kocaman2021biomedical,leaman2008banner}. Different from NER in the general domain, biomedical NER suffers from (1) high semantic and syntactic complexity \cite{kim2022your}, (2) knowledge-intensive nature with ever-growing biomedical concepts \cite{kim2022your}, and (3) acute data scarcity \cite{kang2021umls}. To overcome these, researchers have proposed newer architectures \cite{gridach2017character} or resorted to using knowledge-enhanced, or in-domain pre-trained PLMs \cite{lee2020biobert,10.1093/bib/bbac409}. Contrary to this, data augmentation for low-resource BioNER, or NER in general, is a very under-explored problem in literature.
\vspace{0.5mm}

{\noindent \textbf{Data augmentation for low-resource NER.}} Data augmentation for low-resource NLP is a widely studied problem \cite{feng2021survey}. Most of these systems, when applied to the task of NER, suffer from token-label misalignment, thus demanding specially designed data augmentation techniques. Data augmentation for low-resource NER is a relatively under-studied problem, with systems built on simple word-level modifications like entity replacement \cite{dai2020analysis} or synonym replacement \cite{dai2020analysis} or sophisticated neural learning techniques like LSTM-based Language Modeling \cite{ding2020daga}, Masked Language Modeling with PLMs \cite{zhou-etal-2022-melm}, or auto-regressive language modeling using PLMs \cite{liu2021mulda}. Systems built specifically for BioNER include \cite{phan2022simple}, which modifies the original entity replacement method from \cite{dai2020analysis} and replaces entities of the same type from sentences that are semantically similar to the original sentence. On similar lines, UMLS-EDA \cite{kang2021umls} proposes to replace a UMLS concept from the original sentence and replace it with a randomly selected synonym from UMLS \cite{lindberg1993unified}. Though these systems perform well on benchmark datasets, we acknowledge that none of these systems explicitly focus on the factuality and diversity measures of generated augmentations.
\section{Methodology}
\label{sec:method}
In this section, we give an overview of our proposed BioAug fine-tuning and generation approach. Fig. \ref{fig:BioAug} shows a pictorial representation of the entire workflow. A sentence from our training dataset is first passed through a series of pre-processing steps, and the sentence is corrupted and augmented with external factual knowledge for BioAug fine-tuning. Next, for generating augmentations, the training samples are corrupted similarly to fine-tuning, and the fine-tuned BioAug is conditioned on the corrupted text to generate new augmentations. These generated augmentations are finally added to the training data, followed by fine-tuning a simple transformer model for token-level NER.

\subsection{Training BioAug}
\label{sub:finetuning}


{\noindent \textbf{(1) Keyword Extraction.}} For each sentence in our training dataset with gold NE annotations, we first extract \emph{keywords} from our sentence that provides contextually relevant knowledge about the target NE (e.g., in Figure \ref{fig:BioAug} "mutation" and "frequency"  are important keywords that define the NEs HNPCC and Colorectal Cancer). To achieve this, we use a pre-trained entity extraction model to extract all possible entities from the sentence. Note that for our work, entities differ from the actual annotated NEs - we define entities as domain-specific phrases in the sentence with one or multiple tokens that are not NEs but deemed to be \emph{important} in the context of the sentence. Extracting these entities ensures that while corrupting our sentence (described in later paragraphs), we retain words that help a model understand the true word sense of linguistically complex biomedical NEs. We use SciSpaCy \cite{neumann-etal-2019-scispacy} as our pre-trained entity extraction model.
\vspace{0.5mm}

{\noindent \textbf{(2) Relation Extraction as Additional External Knowledge.}} As discussed earlier, knowledge-intensive tasks like BioNER benefit greatly from storing factual knowledge in their parameters. Though retaining extra entities in a sentence around the NE for text denoising provide additional local context to the model, to provide better global context and enhance its biomedical language understanding capabilities, we add knowledge facts to the corrupted sentence during fine-tuning in the form of entity-entity, entity-NE and NE-NE relation triplets. Augmenting additional external contextual knowledge for text denoising makes the model use and learn this knowledge during training, which in turn prevents the model from hallucinating \cite{10.5555/3524938.3525306,wold2022effectiveness,zhang-etal-2021-smedbert} thus allowing it to generate more factual augmentations. Our knowledge augmentation procedure has similarities with knowledge-augmented LM training systems in literature \cite{zhang-etal-2021-smedbert,kaur-etal-2022-lm} but differs in one key area - unlike prior systems in literature that use knowledge graph neighbors of entity mentions in the sentence as knowledge triples, we use domain-specific biomedical relations between entities as knowledge triples (example in Fig. \ref{fig:BioAug}). Doing this is better suited for our denoising task and allows us to filter knowledge irrelevant to the sentence. To extract these relations, we train a simple relation extraction model \cite{han-etal-2019-opennre} on a large-scale biomedical relation extraction dataset built in an unsupervised manner using UMLS Metathesaurus knowledge graph \cite{bunescu2007learning} on PubMed abstracts \cite{mcentyre2002ncbi}. For more details on building the RE dataset, we refer our readers to \cite{hogan2021abstractified}.
\vspace{0.5mm}

{\noindent \textbf{(3) Dynamic \& Selective Masking.}} Out of all the entities in a sentence identified by SciSpaCy, we first remove the entities that overlap with the original NE and randomly select \emph{e\%} of the remaining entities which we call keywords. \emph{e} is sampled from a from a Gaussian distribution $\mathcal{N}(\mu,\,\sigma^{2})$, where the Gaussian variance $\sigma$ is set to 1/$k$ and $k$ is the total number of entities identified by SciSpaCy. Next, we mask all other tokens in the sentence except the NEs and the identified entities and merge consecutive mask tokens. 
\vspace{0.5mm}

{\noindent \textbf{(4) Sequence Linearization.}} Post selective masking, inspired by various works in literature \cite{zhou-etal-2022-melm,ding2020daga}, we perform sequence linearization and add label tokens before and after each NE token and treat that as the normal context in the sentence. This helps the model to explicitly take label information into consideration.


{\noindent \textbf{(5) Knowledge Augmentation.}} Finally, only knowledge triples belonging to NE and entities left in the sentence are verbalized and concatenated to the masked and linearized sentence from the previous step. This leaves us with a total of ${n \choose 2}$ relations where $n$ is the total number of NEs and keywords in the sentence.
\vspace{0.5mm}

{\noindent \textbf{(6) Fine-tuning.}} For fine-tuning BioAug, we solve a text reconstruction or denoising task using BioBART, where given a corrupted sentence, BioBART learns to recover the entire linearized sentence.
\subsection{Augmentation Generation}
Post fine-tuning BioAug on our novel text reconstruction task, we feed a corrupted sentence to BioAug to generate augmentation. We repeat this for $n$ times to generate $n\times$ augmented samples and finally add this data to the training dataset for fine-tuning our final token classification NER model. Since we generate augmentations auto-regressively, we do \emph{top-k} random sampling with beam search to boost generation diversity. Note that we do not add external knowledge to the corrupted sentence in this step.

\section{Experiments and Results}
\subsection{Datasets}
To prove the efficacy of BioAug, we experiment on 5 biomedical NER datasets, namely BC2GM \cite{krallinger2015chemdner}, BC5DR \cite{li2016biocreative}, NCBI \cite{dougan2014ncbi}, EBMNLP \cite{nye2018corpus}, and JNLPBA \cite{collier2004introduction}. BC2GM is annotated with gene mentions and has 15197/3061/6325 samples in train/dev/test sets, respectively. BC5DR is annotated with chemical and disease mentions and has 5228/5330/5865 samples in train/dev/test sets, respectively. NCBI is annotated with disease mentions and has 5432/787/960 samples in train/dev/test sets, respectively. EBMNLP is annotated with participants, interventions, comparisons, and outcomes spans and has 35005/10123/6193 samples in train/dev/test sets, respectively. JNLPBA is annotated with chemical mentions and has 46750/4551/8662 samples in train/dev/test sets, respectively. For low-resource NER, we sample 100, 200, and 500 sentences from the training dataset and downsample the dev dataset proportionately. We evaluate our NER models on the entire test dataset for all our experiments.


\subsection{Experimental Setup and Baselines}
{\noindent \textbf{Experimental Setup.}} For BioAug, we use BioBART-large \cite{yuan-etal-2022-biobart} as our auto-regressive PLM. We fine-tune BioAug for 10 epochs with a batch size of 16 using Adam optimizer \cite{kingma2014adam} with a learning rate of 1$e^5$. For NER, we use BioBERT-large \cite{lee2020biobert} with a linear head and follow the token-level classification paradigm using the flair framework \cite{akbik2019flair}. We train all our models for 100 epochs with a batch size of 16 using Adam optimizer with a learning rate of 1$e^2$.
\vspace{0.5mm}

{\noindent \textbf{Baselines.}} \textbf{Gold-only} is NER trained on only gold-annotated data from the respective datasets and low-resource splits without any added augmentation. \textbf{SR-UMLS \cite{kang2021umls}} or Synonym Replacement with UMLS, identifies all the UMLS concepts in the sentence and randomly replaces one concept with a randomly selected synonym from UMLS. \textbf{DAGA \cite{ding2020daga}} or Data Augmentation with a Generation Approach, trains a single-layer LSTM-based recurrent neural network language model (RNNLM) by optimizing for next-token prediction with linearized sentences. At the generation step, DAGA uses random sampling to generate entirely new sentences auto-regressively. \textbf{MulDA \cite{liu2021mulda}} or Multilingual Data Augmentation Framework, is trained on a similar objective to DAGA but using BART instead of RNNLM. \textbf{MELM \cite{zhou-etal-2022-melm}} or Masked Entity Language Modeling, fine-tunes a transformer-encoder-based PLM using masked language modeling on linearized sentences. MELM outperforms all other prior art on low-resource settings on the CoNLL 2003 dataset.



\subsection{Results and Analysis}
{\noindent \textbf{Quantitative Analysis.}} Table \ref{tab:quant_results} reports the average micro-averaged F1 over 3 runs with 3 different random seeds for BioAug and all our baselines. As clearly evident, BioAug outperforms all our baselines and achieves absolute improvement in the range of 1.5\%-21.5\%. PLM-based baselines (DAGA, MulDA and MELM) under-perform across all settings, which we attribute to the lack of inadequate knowledge for generating factual augmentations. On the contrary, though PLM-based, BioAug generates effective augmentations due to the local and global context provided to it during fine-tuning. Table \ref{tab:quant_aug} compares generated augmentations on the quantitative measures of perplexity and diversity. Perplexity \cite{jelinek1977perplexity}, calculated using BioGPT \cite{10.1093/bib/bbac409}, can also be seen as a measure of factuality. We calculate two measures of diversity - the average absolute length difference between actual and generated samples (Diversity-L) and the average percentage of new tokens in augmentations (Diversity). BioAug outperforms all prior art in all these metrics.

{\noindent \textbf{Qualitative Analysis.}} Fig. \ref{fig:conll} shows an example of augmentations generated from our baselines and BioAug. As clearly evident, BioAug generated augmentations that are both more factual and diverse.
\begin{table}[t]
\caption{\small Result comparison on 5 benchmark BioNER datasets across 4 dataset size settings. BioAug outperforms all our baselines.}
\tiny
\resizebox{0.90\columnwidth}{!}{
\begin{tabular}{clcccccc}
\hline
\hline
\textbf{\#Size} & \textbf{Model} & \textbf{BC2GM} & \textbf{BC5DR} & \textbf{NCBI} & \textbf{EBMNLP} & \textbf{JNLPBA} & \textbf{Avg.} \\ \hline
                & Gold Only      & 56.94          & 74.90          & 72.99         & 18.81           & 44.37   & 53.60       \\
                & DAGA           & 38.63 & 60.96 & 58.26 & 17.48 & 43.85 & 43.84    \\
                & MulDA           & 39.67 & 62.35 & 59.56 & 20.32 & 45.66 & 45.51     \\
100             & SR-UMLS           & 54.83 & 75.64 & 68.35 & 21.68 & 55.66 & 55.23    \\
                & MELM           & 48.56          & 74.70 & 65.74 & 24.64 & 50.32 & 52.79    \\
                & \textbf{BioAug \emph{(ours)}} & \cellcolor{cyan!20}\textbf{60.17} & \cellcolor{cyan!20}\textbf{77.58} & \cellcolor{cyan!20}\textbf{75.14} &  \cellcolor{cyan!20}\textbf{27.35} & \cellcolor{cyan!20}\textbf{60.00} & \cellcolor{cyan!20}\textbf{60.05} \\ 
\hline
                & Gold Only      & 62.16          & 76.08          & 76.02         & 23.96           & 54.26 & 58.50         \\
                & DAGA           & 48.95          & 68.69          & 70.92         & 23.53           & 53.58 & 53.13          \\
                & MulDA          & 50.11          & 69.35          & 72.28         & 25.37           & 55.28 & 54.48     \\
200             & SR-UMLS       & 62.88          & 78.18          & 74.43         & 27.14           & 63.59 &  61.24         \\
                & MELM           & 58.78          & 79.06          & 73.49         & 21.19           & 58.18 & 58.14          \\
                & \textbf{BioAug \emph{(ours)}}  & \cellcolor{cyan!20}\textbf{67.17} & \cellcolor{cyan!20}\textbf{80.30}  & \cellcolor{cyan!20}\textbf{78.33}  & \cellcolor{cyan!20}\textbf{29.66} & \cellcolor{cyan!20}\textbf{65.40} & \cellcolor{cyan!20}\textbf{64.17} \\ 
\hline
                & Gold Only      & 65.97          & 82.55          & 80.18         & 31.48           & 62.04 & 64.44           \\
                & DAGA           & 53.95          & 76.60          & 78.70         & 32.41           & 61.72 & 60.68         \\
                & MulDA           & 54.92          & 78.04          & 79.92         & 33.53           & 62.63 & 61.81         \\
500             & SR-UMLS           & 65.43          & 82.70          & 79.16        & 32.92       & 65.36 & 65.11          \\
                & MELM           & 58.78          & 81.19          & 75.49         & 32.26         & 61.64  & 61.87        \\
                & \textbf{BioAug \emph{(ours)}} & \cellcolor{cyan!20}\textbf{70.61} & \cellcolor{cyan!20}\textbf{84.48} & \cellcolor{cyan!20}\textbf{80.64} & \cellcolor{cyan!20}\textbf{37.94} & \cellcolor{cyan!20}\textbf{68.07} & \cellcolor{cyan!20}\textbf{68.35}  \\ 
\hline
                & Gold Only      & 82.33          & 89.01          & 87.33         & 42.98           & 74.36 & 75.20          \\
                & DAGA           & 79.62          & 86.69          & 85.15         & 42.46           & 72.52 & 73.29          \\
                & MulDA           & 80.21          & 87.55          & 86.93         & 44.54           & 73.78 & 74.60          \\
All             & SR-UMLS           & 82.18          & 88.48          & 84.66         & 45.75      & 74.93 & 75.20         \\
                & MELM           & 81.46          & 89.18          & 83.95         & 40.38          & 73.82 & 73.76         \\
                & \textbf{BioAug \emph{(ours)}}  &  \cellcolor{cyan!20}\textbf{83.83}  & \cellcolor{cyan!20}\textbf{89.33} & \cellcolor{cyan!20}\textbf{88.14} & \cellcolor{cyan!20}\textbf{47.26} & \cellcolor{cyan!20}\textbf{75.49} & \cellcolor{cyan!20}\textbf{76.81}             \\ 
\hline
\end{tabular}
}
\label{tab:quant_results}
\end{table}

\begin{table}[h]
\caption{\small Quantitative evaluation of generation quality from various systems on the measures of perplexity and diversity.}
    \resizebox{0.90\columnwidth}{!}{
    \tiny
    \centering
        \begin{tabular}{clccc}
        \hline\hline
            \textbf{\#Gold} & \textbf{Method} & \textbf{Perplexity($\downarrow$)} & \textbf{Diversity($\uparrow$)} & \textbf{Diversity-L($\uparrow$)}\\
            \hline
                 & SR-UMLS & 115.76 & 14.65 & 2.38\\
            100      & MELM & 110.50 & 15.83 & 0.0\\
                & \textbf{BioAug \emph{(ours)}} & \cellcolor{cyan!20}\textbf{39.69} & \cellcolor{cyan!20}\textbf{47.88} & \cellcolor{cyan!20}\textbf{9.074}\\ 
             \hline
             & SR-UMLS & 110.23 & 15.33 & 2.56\\
            200 & MELM & 97.78 & 18.65 & 0.0\\
            & \textbf{BioAug \emph{(ours)}} & \cellcolor{cyan!20}\textbf{32.45} & \cellcolor{cyan!20}\textbf{45.67} & \cellcolor{cyan!20}\textbf{9.67}\\
             \hline
             & SR-UMLS & 102.55 & 14.98 & 2.42\\
            500 & MELM & 94.65 & 14.87 & 0.0\\
            & \textbf{BioAug \emph{(ours)}} & \cellcolor{cyan!20}\textbf{31.14} & \cellcolor{cyan!20}\textbf{44.72} & \cellcolor{cyan!20}\textbf{10.17}\\
             \hline
        \end{tabular}
    \label{tab:quant_aug}
    }
\end{table}

\section{Conclusion and limitations}

In this paper, we propose BioAug, a novel data augmentation framework for low-resource BioNER. BioNER is fine-tuned on a novel text reconstruction task and is able to generate diverse and factual augmentations for BioNER. We empirically show that augmentations generated by BioAug prove to be extremely effective for low-resource BioNER. As part of future work we would tackle BioAug's limitation of not introducing new NEs in augmentations.


\bibliographystyle{ACM-Reference-Format}
\bibliography{sample-base}
\end{document}